\documentclass[10pt,twocolumn]{article}

\usepackage{iccv}
\usepackage{times}
\usepackage{epsfig}
\usepackage{graphicx}
\usepackage{amsmath}
\usepackage{amssymb}
\usepackage[utf8]{inputenc} % allow utf-8 input
\usepackage{color}
\usepackage[table,xcdraw]{xcolor}
\usepackage{booktabs}
\definecolor{mypink3}{cmyk}{0, 0.7808, 0.4429, 0.1412/}
\definecolor{myblue3}{cmyk}{0, 0.1418, 0.9412, 0.3412/}

\DeclareMathOperator*{\argmax}{arg\,max}
\DeclareMathOperator*{\argmin}{arg\,min}
\usepackage[linesnumbered,ruled,vlined]{algorithm2e}

\usepackage{multirow}
\usepackage{colortbl}

\newcommand{\figref}[1]{Fig.~\ref{#1}}
\newcommand{\tabref}[1]{Table.~\ref{#1}}
\newcommand{\eqnref}[1]{Eq.~(\ref{#1})}
\newcommand{\algref}[1]{Alg.~(\ref{#1})}

\usepackage[pagebackref=true,breaklinks=true,colorlinks,bookmarks=false]{hyperref}

\iccvfinalcopy % *** Uncomment this line for the final submission

\def\iccvPaperID{7219} % *** Enter the ICCV Paper ID here

\ificcvfinal\fi

\begin{document}

%%%%%%%%% TITLE
 \title{LabOR: Labeling Only if Required for Domain Adaptive Semantic Segmentation}
%\title{LabOR: Labeling Only if Required via Adaptive Pixel Selector guided by Unsupervised Domain Adaptation Model}

\author{%
  Inkyu Shin\quad%
  Dong-Jin Kim\quad
  Jae Won Cho\quad
  Sanghyun Woo\quad%
  Kwanyong Park\quad%
  In So Kweon
  %
  %$^1$KAIST, Daejeon, South Korea.\quad%
  %$^2$Microsoft Research Asia, Beijing, China.\\
  %{\footnotesize{$^1$\texttt{\{djnjusa,jinsc37,iskweon77\}@kaist.ac.kr } \quad %$^2$\texttt{\{xias,stevelin\}@microsoft.com} }}
  
% \authorrunning{I. Shin et al.}
% \titlerunning{Two-phase Pseudo Label Densification}
\\
KAIST, South Korea. \\
\texttt{\{dlsrbgg33,djnjusa,chojw,shwoo93,pkyong7,iskweon77\}@kaist.ac.kr}
\
}

\maketitle

% ABSTRACT
\begin{abstract}
Unsupervised Domain Adaptation (UDA) for semantic segmentation has been actively studied to mitigate the domain gap between label-rich source data and unlabeled target data. Despite these efforts, UDA 
% has a long way to go to reach fully supervised performance. 
still has a long way to go to reach the fully supervised performance.
%In order to mitigate this issue, 
To this end, we propose a \textbf{Lab}eling \textbf{O}nly if \textbf{R}equired strategy, \textbf{LabOR}, where we introduce a human-in-the-loop 
%method 
approach to adaptively give scarce labels to points that a UDA model is uncertain about.
In order to find the uncertain points, we generate an inconsistency mask using %maximum classifier discrepancy~\cite{saito2018maximum}, 
the proposed adaptive pixel selector
and we label these segment-based regions to achieve near supervised performance with only a small fraction (about 2.2\%) ground truth points, which we call ``Segment based Pixel-Labeling (SPL).''
To further reduce the efforts of the human annotator, we also propose ``Point based Pixel-Labeling (PPL),'' which finds the most representative points for labeling within the generated inconsistency mask. This reduces efforts from 2.2\% segment label $\rightarrow$ 40 points label while minimizing performance degradation. 
Through extensive experimentation, we show the 
advantages of this new framework for domain adaptive semantic segmentation  while minimizing human labor costs.

% This proposed 'Adaptive point labeling' outperforms random point selection by 5\% mIoU through top-down methods and at the same time lags only 4\% mIoU behind the full supervision on GTA5 $\rightarrow$ Cityscapes.
\end{abstract}

% Introduction
\section{Introduction}

Semantic segmentation enables understanding of image scenes at the pixel level, and is critical for various real-world applications such as autonomous driving~\cite{Richter_2016_ECCV} or simulated learning for robots~\cite{sim2robot}. 
%But 
Unfortunately, the pixel level understanding task in deep learning requires tremendous 
%labor cost 
labeling efforts
in both time and cost. Therefore, unsupervised domain adaptation (UDA)~\cite{gopalan2011domain} addresses this problem by utilizing and transferring the knowledge of label-rich data (source data) to unlabeled data (target data), which can reduce the labeling cost dramatically~\cite{richter2017playing}. According to the adaptation methodology, UDA can be largely divided into \textbf{\textit{Adversarial learning based}}~\cite{li2019bidirectional,tsai2018learning, vu2019advent, wang2020differential} DA  and \textbf{\textit{Self-training based}}~\cite{mei2020instance, pan2020unsupervised, shin2020two, zou2018domain, zou2019confidence} DA. While the former focuses on minimizing 
task-specific loss for source domain and domain adversarial loss, the self-training strategy retrains the model with generated target-specific pseudo labels. 
% Recent works have boosted the performance by mixing these two methods~\cite{li2019bidirectional,mei2020instance, pan2020unsupervised, shin2020two}. 
\begin{figure}[t]
    \centering 
    \includegraphics[width=0.49\textwidth]{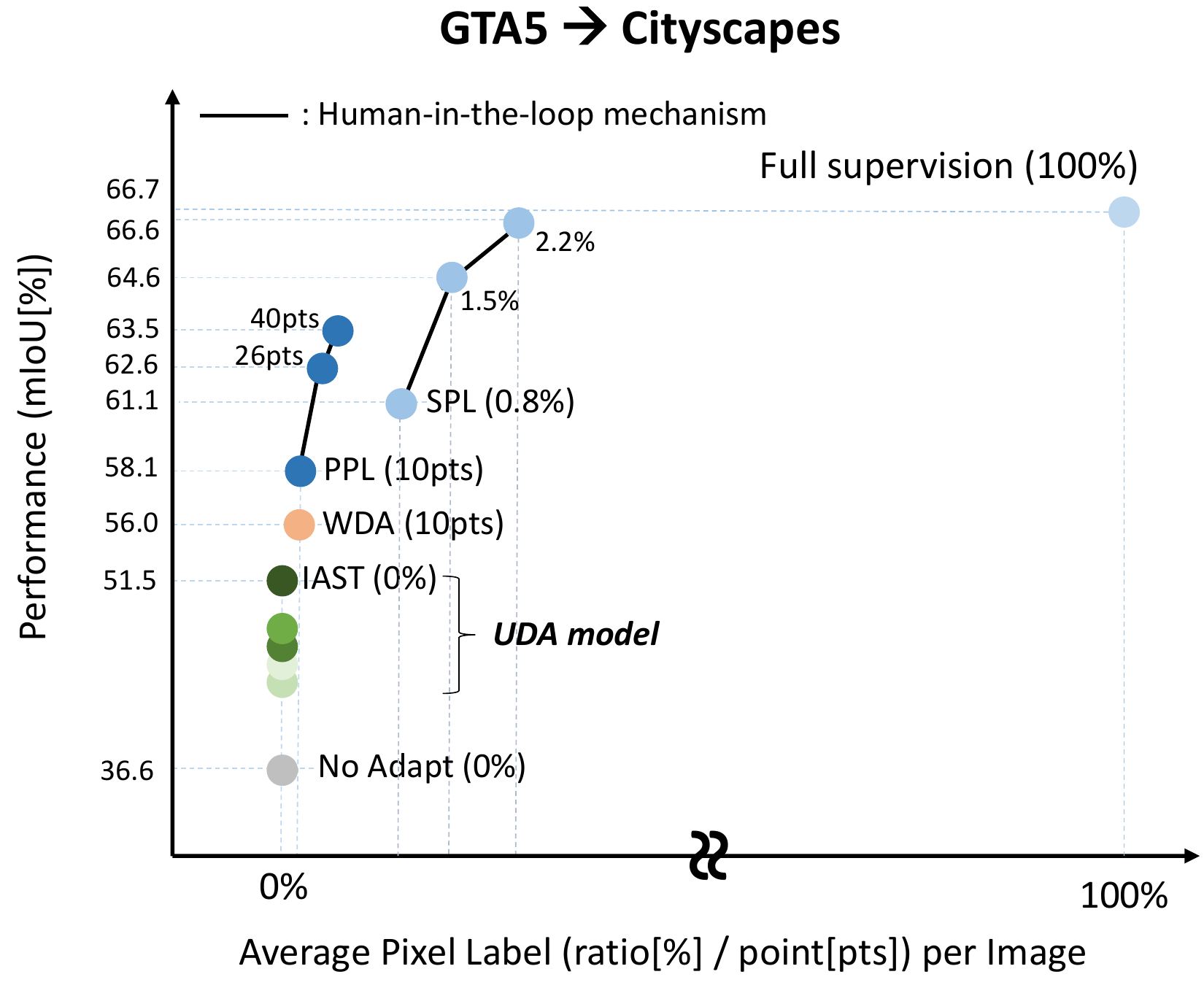}
    \caption{\textbf{Average Pixel Label per image vs. Performance.} Our novel human-in-the-loop framework, \textbf{LabOR} (PPL and SPL) significantly outperforms not only previous UDA state-of-the-art models (e.g., IAST~\cite{mei2020instance}) but also DA model with few labels(e.g., WDA~\cite{Paul_WeakSegDA_ECCV20}). Note that our PPL requires negligible number of label to achieve such performance improvements (25 labeled points per image), and our SPL shows the performance comparable with fully supervised learning (0.1\% mIoU gap).
    Detailed performance can be found in~\tabref{table:gta5_activ} and~\figref{fig:loop_active}.}
    \label{fig:label_ratio_perf}
    \vspace{-3mm}
\end{figure}
Among them, IAST~\cite{mei2020instance} achieves state-of-the-art performance in UDA by effectively mixing adversarial based and self-training based strategies.
% borrows the conventional adversarial based domain adaptation model and proceeds with generating instance adaptive pseudo label.
% initially narrows the distribution gap between the source and target domains through adversarial learning, called the warm-up phase, and then proceeds with self-training methods. 

Despite the relentless efforts in developing UDA models, the performance limitations are clear as it still lags far behind the fully supervision model.
As visualized in~\figref{fig:label_ratio_perf}, the recent UDA methods remain at around ($\sim$50\% mIoU) which is far below the performance of full supervision ($\sim$65\% mIoU) on GTA5~\cite{Richter_2016_ECCV} $\rightarrow$ Cityscapes~\cite{Cordts2016Cityscapes}.
% the performance of recent UDA papers hovers around 50\% which is far below than full supervision model's 65\% mIoU performance on GTA5~\cite{Richter_2016_ECCV} $\rightarrow$ Cityscapes~\cite{Cordts2016Cityscapes}.
% that of the full supervision model, about 65\% mIoU on GTA5~\cite{Richter_2016_ECCV} $\rightarrow$ Cityscapes~\cite{Cordts2016Cityscapes}.

Motivated by the limitation of UDA, we present a new perspective of domain adaptation by utilizing a minute portion of pixel-level labels in an adaptive human-in-the-loop manner. 
%The overview is shown in~\figref{fig:main_figure}.
We name this framework \textbf{Lab}ling \textbf{O}nly if \textbf{R}equired ({LabOR}), which is described in~\figref{fig:main_figure}. 
%On top of the adversarial warm up stage~\cite{} \djkim{(IAST?)}, a model brings out predictions which are known to be pseudo labels. \jw{(what is going on here ???)}
% In order to reduce the initial gap between the source and target domains, introduced as warm-up stage in IAST~\cite{mei2020instance}, we borrow the adversarial based unsupervised model and generate pseudo labels \djkim{(Warmup might be too much detail for Sec.Intro)}\jw{(yes i agree. but it is just one sentence, we can just move to method?)}\djkim{(let's move this sentence to Sec.Method)}. 
Unlike conventional self-training based UDA that retrains the target network with 
%those generated pseudo labels
the pseudo labels generated from the model predictions,
% similar to \cite{saito2018maximum}, 
%we utilize them to \djkim{(utilize the pseudo labels to?)}
%%supervise-train 
%train
%two different classifiers in a supervised manner while maximizing the discrepancy between them \djkim{(We shall clarify this sentence later)}.
we utilize the model predictions to find uncertain regions that require human annotations and train these regions with ground truth labels in a supervised manner.
% so that the uncertain regions can be trained with the ground truth labels in a supervised manner.
In particular, we find regions where the two different classifiers mismatch in predictions.
% This strategy leads to the diversification of classifier models
%This optimization step maximizes the discrepancy between the two classifiers.
In order to effectively find the mismatched regions, we introduce additional optimization step to maximize the discrepancy between the two classifiers like~\cite{cho2021mcdal,saito2018maximum}.
% train two different classifiers supervised by generated target-specific labels while maximizing the discrepancy between them.
Therefore, by comparing the respective predictions from the two classifiers on a pixel level, we create a mismatched area that we call the \emph{inconsistency mask} which can be regarded uncertain pixels.
% With this adaptive pixel selecting model, 
We call this framework the ``Adaptive Pixel Selector'' which guides a human annotator to label on proposed pixels.
% Then, a human annotator intervenes to label on the proposed pixel-position. 
This results in the use of a very small number of pixel-level labels to maximize performance. Depending on how we label the proposed areas, we
propose two different labeling strategies, namely
``Segment based Pixel-Labeling (SPL)'' and ``Point based Pixel-Labeling (PPL).'' While SPL labels every pixels on the inconsistency mask in a segment-like manner, PPL places its focus more on the labeling effort efficiency by finding the representative \emph{points} within a proposed segment.
% We call this whole procedure as `Segment based Pixel Selector (SPS)'. At the same time, we additionally design `Point based Pixel Selector (PPS)' that places its focus more on the labeling effort efficiency by finding the representative \emph{points} within a proposed segment.
We empirically show that the two proposed ``Pixel-Labeling'' options not only help a model achieve near supervised performance but also reduces human labeling costs dramatically.

% To address this challenging issue, we present a new framework to utilize a very small number of labels adaptively while maintaining the ideology of `Minimizing supervision while Maximizing performance'. Gaining inspiration from active learning, we propose to include a human-in-the-loop like oracle to give ground truth labels to certain points that the UDA model deems uncertain. 
% We first use an unsupervised domain adaptation model to generate pseudo label, not to retrain the model but solely for the purposes of guiding the annotator. 
% Then, with these newly annotated pixels, we retrain our model and measure its performance. Following the works of self-training DA models, we use a 3 stage approach and repeat the above steps in 3 stages. We show through our experimentation the effectiveness of using a very minute number of ground truth points to increase the performance of DA to that of fully supervised training.

We summarize our contributions as follows: 
\begin{enumerate}
\vspace{-1mm}
\setlength\itemsep{0.3em}
    \item We design a new framework of domain adaptation for semantic segmentation, LabOR, by utilizing a small fraction of pixel-level labels with an adaptive human-in-the-loop pixel selector.   
    \item We propose two labeling options, Segment based Pixel-Labeling (SPL) and Point based Pixel-Labeling (PPL), and show that %both of these methods have merits in their own rights.
    these methods are especially advantageous in performance compared to UDA and labeling efficiency respectively. 
    % (\ik{ik: I think we need to clarify like ``performance compared to UDA'' and `` efficiency compared to sup''})
    \item We conduct extensive experiments to show that our model outperforms previous UDA model by a significant margin even with very few pixel-level labels. 
\end{enumerate}

% Related Work
\section{Related Work}

\noindent\textbf{Unsupervised Domain Adaptation.} 
Domain Adaptation is a classic computer vision problem that aims to mitigate the performance degradation due to a distribution mismatch across domains and has been investigated in image classification problems through both conventional methods~\cite{fernando2013unsupervised,gong2012geodesic,gopalan2011domain,kulis2011you,li2017domain} and deep CNN-based methods~\cite{ganin2014unsupervised,ghifary2016deep,li2017deeper,long2015learning,motiian2017unified,panareda2017open,sener2016learning}.
Domain adaptation has recently been studied in other vision tasks such as object detection~\cite{chen2018domain}, depth estimation~\cite{atapour2018real}, and semantic segmentation~\cite{pmlr-v80-hoffman18a}.
% In this work, we are particularly interested in \textit{unsupervised} domain adaptation for the task of semantic segmentation.
With the introduction of the automatically annotated GTA dataset~\cite{richter2017playing}, \textit{unsupervised} domain adaptation (UDA) for semantic segmentation has been extensively studied.
Adversarial learning approaches have aimed to minimize discrepancy between source and target feature distributions and this approach has been studied on three different levels in practice:
input-level alignment~\cite{Chen_2019_CVPR, pmlr-v80-hoffman18a,imagetrans,learnSi}, intermediate feature-level alignment~\cite{fcnwild, Hong_2018_CVPR, trasdeepadat, Yawei2019Taking, vu2019advent}, and output-level alignment~\cite{tsai2018learning}.

\noindent\textbf{Domain Adaptation with Few Labels.}
Despite extensive studies in UDA, the performance of UDA is known to be much lower than that of supervised learning~\cite{saito2019semisupervised}.
In order to mitigate this limitation, various works have tried to leverage ground truth labels for the target dataset.
For example, semi-supervised domain adaptation, which utilize randomly selected image-level labels per class as the labeled training target examples, has been recently studied for image classification~\cite{saito2019semisupervised}, semantic segmentation~\cite{wang2020alleviating}, and image captioning~\cite{chen2017show,kim2019image}.
However, these naive semi-supervised learning approaches do not consider which target images should be labeled given a fixed budget size.
Similar to semi-supervised domain adaptation, some works have used active learning~\cite{settles2012active} to give labels to a small portion of the dataset~\cite{su2020active, prabhu2020active}. These works leverage a model to find data points that would increase the performance of the model the most. 
Furthermore, in order to reduce the labeling effort per image for target images in domain adaptation, a method to leverage weak labels, several points per image, has also been studied~\cite{Paul_WeakSegDA_ECCV20}.

In contrast, our work differentiates itself by allowing the model to automatically pinpoint to the human annotator which points to label on a pixel-level that would have the best potential performance increase instead of randomly picking labels which can possibly be already easy for the model to predict. In addition, unlike the semi-supervised model which has random annotations prior to training, we allow the model to let the annotator know which points in an image are best to increase performance.
Although at first glance our method may seem similar to active learning in the human-in-the-loop aspect, our work is the first to propose a method on the \emph{pixel-level} instead of image-level.
Overall, our pixel-level sampling approach is not only efficient, but also orthogonal to the existing active, weak label, or semi-supervised domain adaptation frameworks.

% However, while \cite{Paul_WeakSegDA_ECCV20} make human annotators to decide what points to label, our method algorithmically provide the best points to label, which makes the labelling process similar to that of simple image classification dataset.
% Not only being efficient, but also our method recommends the points that have the best potential performance improvements if labeled, whereas human annotators for \cite{Paul_WeakSegDA_ECCV20} might tend to label easy points that a model might already correctly predict the class.

% Although our method is similar to active learning in the human-in-the-loop aspect, our work is the first to propose a method on the \emph{pixel-level} instead of image-level.
% Therefore, our point-level sampling approach is orthogonal to the existing active or semi-supervised domain adaptation frameworks that utilize image-level sampling approaches.

% % Method
% \input{latex/Categories/03 Problem Definition}

% Method
\section{Proposed Method}

In this section, we introduce our method from inconsistency mask generation to adaptive pixel labeling.

\begin{figure*}[t]
    \centering 
    \includegraphics[width=1\textwidth]{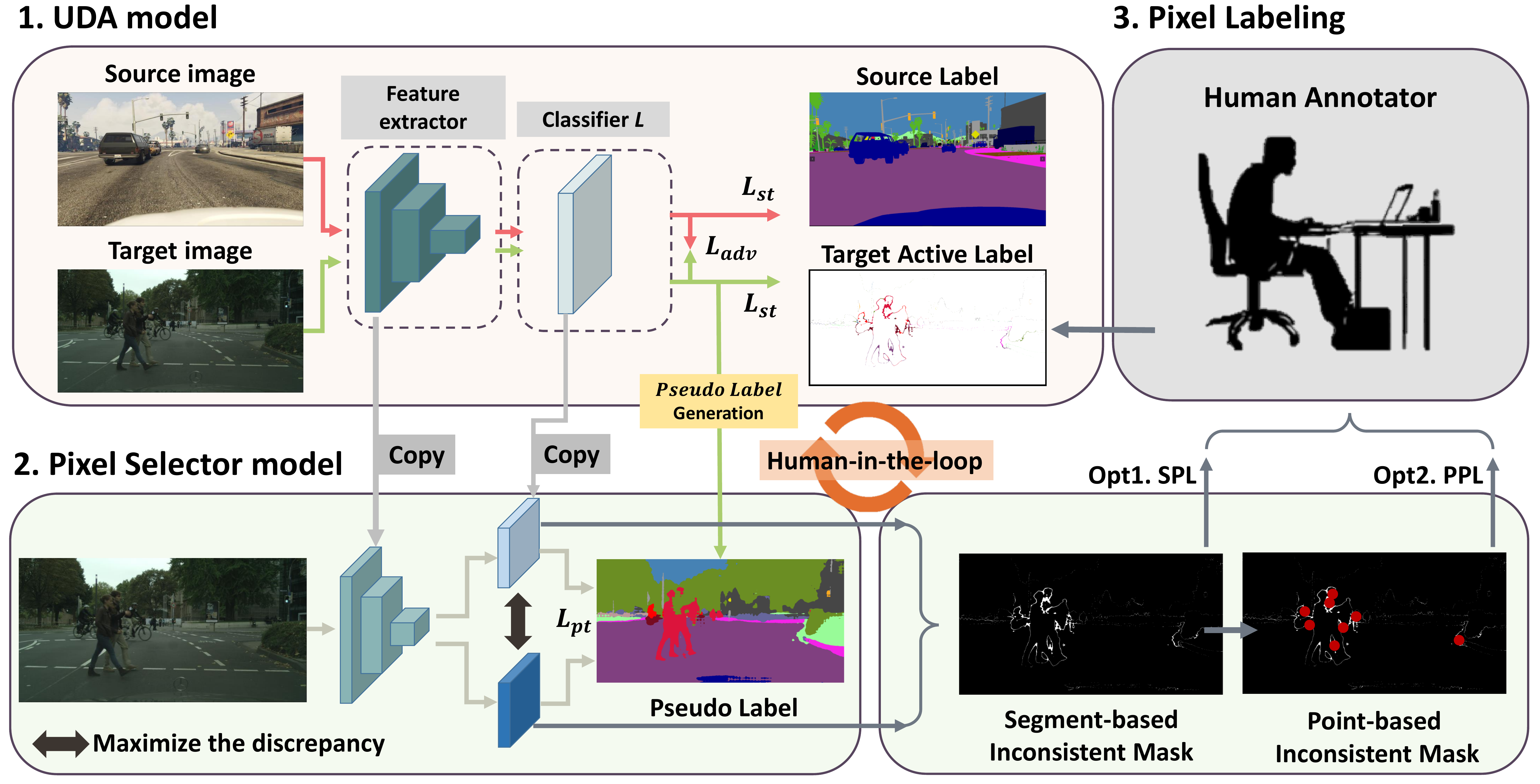}
    \caption{\textbf{The overview of the proposed adaptive pixel-basis labeling, LabOR.} This framework is made up of two models: UDA model and Pixel selector model. The UDA model initially trained from conventional adversarial learning forwards target image to generate pseudo label. Different from normal self-training training scheme~\cite{mei2020instance} that utilizes the generated label to retrain the model directly, we instead train a pixel selector model to brings out inconsistent mask where human annotator is guided to label. In this process, we use pseudo label training loss, $L_{pt}$ which contains pseudo label cross entropy loss and classifiers' discrepancy loss.
    With those human labels, we return to the original UDA model for training that uses $L_{st}$.}
    \label{fig:main_figure}
    \vspace{-3mm}
\end{figure*}

\subsection{Problem Definition: Domain Adaptation}

Let us denote $g_\phi(\cdot)$ as the network backbone with the parameter $\phi$ that generates features from an input $\mathbf{x}$.
Then, with the classification layer including softmax activation $f_\theta(\cdot)$ with the parameter $\theta$, a class prediction (probability) is computed ($\hat{\mathbf{Y}} = p(\mathbf{Y}|\mathbf{x};\theta,\phi)=f_\theta \circ g_\phi(\mathbf{x})\in \mathbb{R}^{W \times H \times K}$, where $W$ and $H$ are width and height of the segmentation map, and $K$ is the total number of classes).
The combined network $f_\theta \circ g_\phi(\cdot)$ can be implemented with typical semantic segmentation generators~\cite{deep2,deep3}. 
A typical semantic segmentation model is trained with cross-entropy loss ${\text{CE}}(\cdot,\cdot)$ with the ground truth label $\mathbf{Y}\in\mathbb{R}^{W \times H}$. Furthermore, let us denote $\mathcal{S}=\{(\mathbf{x}_s, \mathbf{Y}_s)\}_{s=1}^S$ as the labeled images from the source dataset and $\mathcal{T}=\{\mathbf{x}_t\}_{t=1}^{T}$ as the unlabeled images from the target dataset. 
Unsupervised Domain Adaptation (UDA) tries to leverage both the abundant labeled source dataset and the small number of unlabeled target dataset to train a deep neural network.

Recent unsupervised domain adaptive semantic segmentation use self-training methods~\cite{mei2020instance,Zou_2019_ICCV} and have shown state-of-the-art performances and are optimized as follows:
% \begin{equation}
%     \begin{split}
%         \min_{\theta,\phi,\{\tilde{\mathbf{Y}}_t\}} &\mathbb{E}_{(\mathbf{x}_s,\mathbf{Y}_s)\in\mathcal{S}} \bigl[ \mathcal{L}_{\text{CE}}(\mathbf{Y}_s,p(\mathbf{Y}|\mathbf{x}_s;\theta,\phi)) \bigr] \\
%         & + \mathbb{E}_{\mathbf{x}_t\in\mathcal{T}} \bigl[ \mathcal{L}_{\text{self}}(\tilde{\mathbf{Y}}_t,p(\mathbf{Y}|\mathbf{x}_t;\theta,\phi)) \bigr],
%     \end{split}
% \end{equation}
% where $\mathcal{L}_{\text{self}}(\cdot,\cdot)$ is the self-training loss function for the target dataset.
In practice, 
the model alternates between generating pseudo-labels $\tilde{\mathbf{Y}}_t(\mathbf{x}_t)\in\mathbb{R}^{W \times H}$ for an image $\mathbf{x}_t$ based on the model prediction $p(\mathbf{Y}|\mathbf{x};\theta,\phi)$ and retraining the model on the target dataset with the generated pseudo labels. % via $\mathcal{L}_{\text{self}}(\cdot,\cdot)$.
The goal of self-training based domain adaptation~\cite{mei2020instance,Zou_2019_ICCV} is to devise an effective loss function and a way to generate pseudo labels. Specifically, CRST~\cite{Zou_2019_ICCV} propose class-balanced pseudo label generation strategy and confident region KLD minimization to prevent overfitting on pseudo labels. IAST~\cite{mei2020instance} tackles the class-balanced pseudo label generation which ignores the individual attributes of instance to design an instance adaptive selector. Moreover, IAST adds an entropy minimization approach on unlabeled pixels. 
% Even with all efforts in self-training based domain adaptation, it's 
Self-training based domain adaptation far underperforms a fully supervised model. 
This can be attributed to two reasons. First, cutting out unconfident pixels and re-training with the thresholded labels is not intuitive as the model forced to be trained with only the pixels that model itself is confident in. Second, existing pseudo label generation commonly originates from specific manually set hyperparameters, causing incorrect pseudo labels which degrades the performance. 
To address this issue, we propose a new perspective of self-training based domain adaptation with a human-in-the-loop approach by using a human annotator to label a small number of informative \emph{pixels}. 
As the human annotator annotates the pixels where the model is uncertain, the labeled pixels ultimately act as a guide for the model.
We call this method \textbf{Lab}eling \textbf{O}nly if \textbf{R}equired (\textbf{LabOR}).
In order to minimize the efforts of the human annotator, we must answer the key question \emph{``what is an informative pixel to label?''}
In other words, our goal is to find the pixels where the model is uncertain.
To this end, we propose to select the pixels that show the highest \emph{classifier discrepancy} motivated by the classifier discrepancy based domain adaptation method, MCDDA~\cite{saito2018maximum}.
% The detailed process can be found in the next subsection.

%\subsection{Uncertainty Measure with Classifier Discrepancy}
\subsection{Generating Inconsistency Mask}

%In light of the previous section, we propose to include a human-in-the-loop for our method so as the guide our model in parts that it is unsure about. 
%In order to create our uncertain area we use the maximum classifier discrepancy. 
%Borrowing the idea from MCDDA~\cite{saito2018maximum}, instead of using the classifiers to train and use for inference, we simply increase the distance between the classifiers to try to find the points that are still difficult to train and match. 
% Then we use the different outputs from the classifiers and compare them in a pixel-to-pixel manner. 
% In doing so, we simply find the points that do not line up and create a mask that shows the inconsistencies between the two classifiers, we call this the \emph{inconsistency mask}. 
% We assume that this \emph{inconsistency mask} is the points where the model is the most unsure about, and if we give this to a model and guide the model, the model would be able to bridge the gap between the domains. 
% The model is then trained with the given ground truth labeled points and this process is repeated. 
% As we repeat the process, the points that the model is uncertain about will start to decrease, and following the works of self-training, we repeat the steps 3 times.

% The overview of the proposed method is illustrated in 
\figref{fig:main_figure} illustrates an overview of our proposed method.
%First, given a model pre-trained with the source dataset $\mathcal{S}$, we apply typical self-training method~\cite{mei2020instance,zou2018domain} % [CBST, IAST]\jw{(is this the official name? or is there something else we can call it)} 
%to adopt the model on target dataset ($f_\theta \circ g_\phi (x)$) \djkim{(did we? fact check plz)}.
First, we pre-train a model with the labeled source dataset $\mathcal{S}$ by minimizing supervised cross-entropy loss:
\begin{equation}
    %\begin{split}
    %\min_{\theta,\phi} 
    \mathcal{L}_{s}(\theta,\phi)
    =
    %\min_{\theta,\phi} 
    \mathbb{E}_{(\mathbf{x}_s,\mathbf{Y}_s)\in\mathcal{S}} \bigl[
    {\text{CE}}(\mathbf{Y}_s,p(\mathbf{Y}|\mathbf{x}_s;\theta,\phi)) 
    \bigr].
    %\end{split}
\label{eq:source}
\end{equation}
Following this, in order to improve the effectiveness of self-training, 
we utilize warm-up with adversarial training~\cite{mei2020instance} before moving on to self-training.
% Then, we use this classifier $p(\mathbf{Y}|\mathbf{x};\theta,\phi$ to find the uncertain pixels in an image.
\begin{equation}
    %\begin{split}
    %\min_{\theta,\phi} 
    \mathcal{L}_{adv}(\theta,\phi)
    =
    %\min_{\theta,\phi} 
    \mathbb{E}_{\mathbf{x}_s\in\mathcal{S},\mathbf{x}_t\in\mathcal{T}} \bigl[
    {\text{Adv}}(p(\mathbf{x}_s;\theta,\phi), p(\mathbf{x}_t;\theta,\phi)) 
    \bigr].
    %\end{split}
\label{eq:adv}
\end{equation}
Then we copy the parameters of the backbone and the classifier (twice for classifier) (i.e., $\theta^{'}_1 \leftarrow \theta, \theta^{'}_2 \leftarrow \theta, \phi^{'} \leftarrow \phi$) to create our Adaptive Pixel Selector model ($f_{\theta^{'}_1}, f_{\theta^{'}_2}, g_{\phi^{'}}$). This model is only used for the purposes of pixel selection and has no effect on the performance.
Using this newly created model, we optimize the model with the two auxiliary classifiers and increase the discrepancy in relation to each other. After this, we propose to find the pixels where the two classifiers have different output class predictions.
Using the different output class predictions, we create a mask consisting of pixels that are inconsistent $M(\mathbf{x}_t;\phi^{'},\theta^{'}_1,\theta^{'}_2)\in \mathbb{R}^{W \times H}$, and we call this the \emph{inconsistency mask}. 
The mask generation would be formulated as follows:
\begin{equation}
    M(\mathbf{x}_t) = \bigl[ \argmax_K f_{\theta^{'}_1} \circ g_{\phi^{'}} (\mathbf{x}_t)  \neq \argmax_K f_{\theta^{'}_1} \circ g_{\phi^{'}} (\mathbf{x}_t) \bigr] .
\label{eq:inconsistency}
\end{equation}
For simplicity, we abuse the notation $M(\mathbf{x}_t;\phi^{'},\theta^{'}_1,\theta^{'}_2)$ as $M(\mathbf{x}_t)$.
We conjecture that if the two classifiers trained on the same dataset generate different predictions for the same region, then it means the model prediction shows a high variance in that input region. 
% Therefore, we propose to provide more information (labels) to that data points to improve the generalizability of the model.
Therefore we conclude that
this \emph{inconsistency mask} represents the pixels the model is the most unsure about.
In other words, we hypothesize that by giving ground truth labels for these pixels to guide the model, the model would more easily bridge the gap between the domains and improve the generalizability of the model. 
The detailed method on giving ground truth labels will be described in the next subsection.

Given $\phi^{'},\theta^{'}_1, \theta^{'}_2$, we first apply the self-training loss function with the pseudo labels (one-hot vector labels generated from  $\hat{\mathbf{Y}}_t = p(\mathbf{Y}|\mathbf{x}_t;\theta,\phi) $), which has been utilized in various tasks~\cite{kim2018disjoint,kim2020detecting,mei2020instance,sohn2020fixmatch,Zou_2019_ICCV}:
\begin{equation}
    \begin{split}
        &\mathcal{L}_{\text{self}}(\phi^{'},\theta^{'}_1, \theta^{'}_2)\\
        &=\mathbb{E}_{\mathbf{x}_t\in\mathcal{T}} \bigl[  {\text{CE}}( \argmax_K \hat{\mathbf{Y}}_t    ,p(\mathbf{Y}|\mathbf{x}_t;\theta^{'}_1,\phi^{'})) \\
        &\quad\quad\quad +{\text{CE}}( \argmax_K{\hat{\mathbf{Y}}_t}    ,p(\mathbf{Y}|\mathbf{x}_t;\theta^{'}_2,\phi^{'}))
        \Bigr].
    \end{split}
\label{eq:self_training}
\end{equation}
The detailed design choices for the pseudo labels for $\mathcal{L}_{\text{self}}$ is discussed in the supplementary materials. %\tabref{tab:Abl_cut}.
Then, in order to optimize the two auxiliary classifiers to increase the discrepancy in relation to each other, we introduce an additional training stage to optimize the auxiliary classifiers to increase the distance between the classifiers' outputs.
In addition, we also minimize the classifier discrepancy with respect to the backbone feature extractor $g_{\phi^{'}}$, which results in a similar formulation to the classifier discrepancy maximization in MCDDA~\cite{saito2018maximum}:
\begin{equation}
    \begin{split}
        \min_{\phi^{'}}\max_{\theta^{'}_1, \theta^{'}_2}& \mathcal{L}_{\text{dis}}(\phi^{'},\theta^{'}_1, \theta^{'}_2)\\
        =\min_{\phi^{'}}\max_{\theta^{'}_1, \theta^{'}_2}& \mathbb{E}_{\mathbf{x}_t\in\mathcal{T}} 
        \Bigl[ ||f_{\theta^{'}_1} \circ g_{\phi^{'}} (\mathbf{x}_t) - f_{\theta^{'}_2} \circ g_{\phi^{'}} (\mathbf{x}_t)||_1 \Bigr].
    \end{split}
\label{eq:maximization}
\end{equation}
Note that the goal of classifier discrepancy maximization in MCDDA is to create tighter decision boundaries in order to align the latent feature distributions between the source and the target domains.
In contrast, we maximize the classifier discrepancy for the sole purposes of generating a more representative inconsistency mask so that the human annotator can give ground truth labels to pixels that truly require labels.
After optimizing the auxiliary classifiers ($\theta^{'}_1, \theta^{'}_2, \phi^{'}$), we utilize the different outputs from these classifiers and compare them in a pixel-to-pixel manner using~\eqref{eq:inconsistency} to obtain $M(\mathbf{x}_t)$. 
After the human annotator gives ground truth labels to the uncertain pixels based on $M(\mathbf{x}_t)$, the model ($f_\theta, g_\phi$) is then trained with the target dataset $\mathcal{T}$ with the given ground truth labeled pixels $\tilde{\mathbf{Y}}_t(\mathbf{x}_t)$:
\begin{equation}
    %\begin{split}
        %\min_{\theta,\phi} 
        \mathcal{L}_{t}(\theta,\phi)
        =
        %\min_{\theta,\phi} 
        \mathbb{E}_{\mathbf{x}_t\in\mathcal{T}} \bigl[
        {\text{CE}}(\tilde{\mathbf{Y}}_t(\mathbf{x}_t),p(\mathbf{Y}|\mathbf{x}_t;\theta,\phi)) 
        \bigr].
    %\end{split}
\label{eq:target}
\end{equation}
Then the process starting from copying ($\theta^{'}_1 \leftarrow \theta, \theta^{'}_2 \leftarrow \theta, \phi^{'} \leftarrow \phi$), optimizing $\mathcal{L}_{\text{self}}(\phi^{'},\theta^{'}_1, \theta^{'}_2)$ and $\mathcal{L}_{\text{dis}}(\phi^{'},\theta^{'}_1, \theta^{'}_2)$, to inconsistency generation $M(\mathbf{x}_t)$ is repeated. 
The overall method is summarized in \algref{alg:APS}.
We repeat the process 3 times as we empirically found that the number of uncertain pixels and the model performance converges after 3 stages.

% Also, note that the auxiliary layers ($\theta^{'}_1, \theta^{'}_2, \phi^{'}$) are only used to measure the uncertainty, and we train our model ($f_\theta, g_\phi$) again with the parameter $\theta,\phi$ with the newly annotated uncertain pixels.

\begin{algorithm}[ht]
\DontPrintSemicolon
\KwIn{
Source data $\mathcal{S}$, Target data $\mathcal{T}$, Initialized model $f_\theta \circ g_\phi(\cdot)$
}
\KwOut{
The model with adapted weights on target dataset $f_{\theta} \circ g_{\phi}(\cdot)$
} 
\Begin
{
    Pre-train the model on the source dataset.\\
    And, initial adapt with adversarial learning.\\
     $\theta,\phi \leftarrow \argmin_{\theta,\phi} \mathcal{L}_{s}(\theta,\phi)+\mathcal{L}_{adv}(\theta,\phi)$ (\eqnref{eq:source})\\ %\argmin_{\theta,\phi} \mathbb{E}_{(\mathbf{x}_t,\mathbf{Y}_s)\in\mathcal{S}} \bigl[ \mathcal{L}_{\text{CE}}(\mathbf{Y}_s,f_{\theta} \circ g_{\phi} (\mathbf{x}_t)) \bigr] $\\
    %$\theta,\phi \leftarrow \argmin_{\theta,\phi} \mathbb{E}_{(\mathbf{x}_t,\mathbf{Y}_s)\in\mathcal{S}} \bigl[ \mathcal{L}_{\text{CE}}(\mathbf{Y}_s,p(\mathbf{Y}|\mathbf{x}_s;\theta,\phi)) \bigr] $\\

    \For{3 Stages}
        {
            Define auxiliary layers and copy weights\\
            $\theta^{'}_1 \leftarrow \theta$,\quad
            $\theta^{'}_2 \leftarrow \theta$,\quad
            $\phi^{'} \leftarrow \phi$\\
            
            Apply self-training (\eqnref{eq:self_training})\\
            $\phi^{'},\theta^{'}_1,\theta^{'}_2 \leftarrow \displaystyle\argmin_{\phi^{'}, \theta^{'}_1, \theta^{'}_2} \mathcal{L}_{\text{self}}(\phi^{'},\theta^{'}_1, \theta^{'}_2) $ \\
            
            %Optimize the copied layers 
            Maximize classifier discrepancy (\eqnref{eq:maximization})\\
            $\phi^{'},\theta^{'}_1,\theta^{'}_2 \leftarrow \displaystyle\argmin_{\phi^{'}}\max_{\theta^{'}_1, \theta^{'}_2} \mathcal{L}_{\text{dis}}(\phi^{'},\theta^{'}_1, \theta^{'}_2) $ \\  %\mathbb{E}_{\mathbf{x}_t\in\mathcal{T}} \Bigl[ ||f_{\theta^{'}_1} \circ g_{\phi^{'}} (\mathbf{x}_t) - f_{\theta^{'}_2} \circ g_{\phi^{'}} (\mathbf{x}_t)||_1 \Bigr]$\\
            %$\phi^{'},\theta^{'}_1,\theta^{'}_2 \leftarrow \argmin_{\phi^{'}}\max_{\theta^{'}_1, \theta^{'}_2} \mathbb{E}_{\mathbf{x}_t\in\mathcal{T}} \Bigl[ ||p(\mathbf{Y}|\mathbf{x}_t;\theta^{'}_1,\phi^{'}) - p(\mathbf{Y}|\mathbf{x}_t;\theta^{'}_2,\phi^{'})    ||_1 \Bigr]$\\
            
            \For{$\mathbf{x}_t\in\mathcal{T}$}
            {
                Generate $M(\mathbf{x}_t;\phi^{'},\theta^{'}_1,\theta^{'}_2)$ with \eqnref{eq:inconsistency}. \\
                %$M(\mathbf{x}_t) = \bigl[ \argmax_K f_{\theta^{'}_1} \circ g_{\phi^{'}} (\mathbf{x}_t)  \neq \argmax_K f_{\theta^{'}_1} \circ g_{\phi^{'}} (\mathbf{x}_t) \bigr]$\\
                
                %$(\mathbf{x}_t, \mathbf{Y}_t) = \text{ORACLE}(\mathbf{x}_t)$\\
                \If{SPL}
                {
                    Annotate inconsistency mask \\  
                    $\tilde{\mathbf{Y}}_t(\mathbf{x}_t) \leftarrow M(\mathbf{x}_t) \odot \mathbf{Y}_t$\\
                }
                
                \ElseIf{PPL}
                {
                    %y_p || Pick Representative Points\\
                    %Annotate y_p\\
                    %y_gt \leftarrow y_p
                    Select representative points\\
                    $P(\mathbf{x}_t) = \text{SelectPt}(M,p(\mathbf{Y}|\mathbf{x}_t;\theta,\phi))$
                    
                    Annotate the points \\
                    $\tilde{\mathbf{Y}}_t(\mathbf{x}_t) \leftarrow P(\mathbf{x}_t) \odot \mathbf{Y}_t$\\
                    
                }
            
            }

            Train the model with the pseudo labels \\ %f_{\theta} \circ g_\phi (x) \\
            %$\theta,\phi \leftarrow \displaystyle\argmin_{\theta,\phi} \mathbb{E}_{\mathbf{x}_t\in\mathcal{T}} \bigl[ \mathcal{L}_{\text{CE}}(\tilde{\mathbf{Y}}_t(\mathbf{x}_t),f_{\theta} \circ g_{\phi} (\mathbf{x}_t))$
            $\theta,\phi \leftarrow \argmin_{\theta,\phi} \mathcal{L}_{t}(\theta,\phi)$ (\eqnref{eq:target})\\
            %$\theta,\phi \leftarrow \argmin_{\theta,\phi} \mathbb{E}_{\mathbf{x}_t\in\mathcal{T}} \bigl[ \mathcal{L}_{\text{CE}}(\tilde{\mathbf{Y}}_t(\mathbf{x}_t),p(\mathbf{Y}|\mathbf{x}_t;\theta,\phi))$\\

            %\If{Not Stage 3}
            %    {
            %    Train f_{\theta^{'}_1} \circ g_\phi (x)~and~f_{\theta^{'}_2} \circ g_\phi (x)
            %    }
        }
      
}
\caption{Pixel Selector Model}
\label{alg:APS}
\end{algorithm}

%Here, the motivation to optimize the auxiliary classifiers is that we try to find the points that are still difficult to train and match \djkim{(really?)}\jw{(pls check)}. 

\subsection{Adaptive Pixel Labeling}

%However, a question arises as to which points we should label. 
%Of course, the best possible method is to give labels to all points that the model is uncertain about.
%Although this method is very powerful, it requires around 3\% labels, which if we calculate amounts to ~~~~~ many labeled points. 
%Although this is a fraction in comparison to the entire image which requires ~~~ (819,200???) points, we believe that we can further reduce the number of points drastically without sacrificing performance too much. 
Given an inconsistency mask $M(\mathbf{x}_t)$, the question arises as how to give labels to the pixels.
With this in mind, we propose two different methods for giving ground truth annotations with different focuses and strengths.

\noindent\textbf{Segment based Pixel-Labeling (SPL).}
As the inconsistency mask shows all pixels that the model is uncertain about, we consider giving ground truth annotations for all the pixels selected. We call this methods the Segment based Pixel-Labeling (SPL). 
% For SPL, given an inconsistency mask, we propose to label every single point of a given inconsistency mask. 
In SPL, no further calculations are needed after the inconsistency mask has been generated, and after the pixels are annotated, the model $p(\mathbf{Y}|\mathbf{x};\theta,\phi)$ is further trained.
Empirically, we find that the inconsistency mask for each stage averages in percent of pixel of total pixels per image at 1\% and totals to 2.2\% at the final stage as some uncertain pixels are overlapped. The performance of SPL achieves near supervised learning, and it far exceeds the performance of our next method, which is more focused on drastically reducing human annotation labor.

\noindent\textbf{Point based Pixel-Labeling (PPL).}
We also propose another Pixel-Labeling method that sets its focus on minimizing human annotation costs; we call this method the Point based Pixel-Labeling (PPL). 
Although PPL receives an inconsistency mask like SPL, we propose to label only the most \emph{representative} pixels in the inconsistency mask instead of labeling all the pixels. Among the most representative pixels, we deliberately choose to maximize diversity by selecting all unique classes present in the inconsistency mask.
% \djkim{(we shall mention the motivation of our method, including about diversity)}

Given a set of uncertain pixels (inconsistency mask $M(\mathbf{x})$) and a model's output probability prediction for all the pixels  $\hat{\mathbf{Y}}=\{\hat{\mathbf{Y}}_{i,j}\in \mathbb{R}^K | i\in [1,W], j \in [1,H]  \}$, we first cluster the pixels that the model $p(\mathbf{Y}|\mathbf{x};\theta,\phi)$ predicts to be the same class. We define the set of uncertain pixels $\mathcal{D}^{k}$ for class $k$ as follows:
\begin{equation}
    \mathcal{D}^{k} = \{(i,j) \in M(\mathbf{x})| k = \textstyle\argmax_K \hat{\mathbf{Y}}_{i,j}\}.
\end{equation}
Then we compute the class prototype vector ${\mu}^k$ for each class $k$ as the mean vectors of $\mathcal{D}^k$:
\begin{equation}
    {\mu}^k = \frac{1}{|\mathcal{D}^{k}|} \sum_{(i,j)\in \mathcal{D}^{k}} \hat{\mathbf{Y}}_{i,j} 
    \in \mathbb{R}^{K}.
\end{equation}
Finally, we select the points that has the most similar probability vector for each prototype vector to construct the set of selected points $P$ :
\begin{equation}
    P(\mathbf{x}) = \bigl\{\argmin_{(i,j)\in \mathcal{D}^{k}} d \bigl({\mu}^k , \hat{\mathbf{Y}}_{i,j} \bigr) \bigr\}^{K}_{k=1}.
\label{eq:ppleq}
\end{equation}
We use cosine distance for a distance measure $d(\cdot,\cdot)$.
Note that as $\mathcal{D}^k$ can be a null set for some classes, $0\leq |P(\mathbf{x}_t)| \leq K$, if the model fails to predict a certain class. 
At each stage, on average, the model generates 12 clusters, and cumulatively we average on giving 40 ground truth labels per target image $\mathbf{x}_t$ in an image of size $640 \times 1280$. This calculates to a $\approx 0.0049\%$ of the image being given ground truth labels. In comparison to SPL, which averages $\approx 18022$ pixels $\rightarrow 2.2\%$ of entire image, we further reduce the human labeling costs by $0.2\%$.
Due to the drastically reduced amount of ground truth annotations, PPL naturally under-performs in relation to SPL. 
Nevertheless, we empirically show that the performance gain of PPL over other UDA or weakly supervised DA methods is still significant.
%we believe that the benefits of PPL, and the performance gain over other UDA or weakly supervised DA methods are worth noting.

% To further reduce the number of points, we believe that with our given inconsistency mask, we can pick a representative point within the inconsistency mask instead of giving all the points labels. In doing so, we can drastically reduce the number of points from 3\% to around 30 points total. In order to pick the representative points, we use a metric (this metric) and describe our method

% Experiments

\begin{table*}[t]
\renewcommand{\arraystretch}{1.2}
\begin{center}
\centering
\resizebox{\textwidth}{!}{
\begin{tabular}{l c c c c c c c c c c c c c c c c c c c|c}
% \hline
\toprule
\multicolumn{21}{c}{GTA5 $\to$ Cityscapes}\\
\midrule
Method  & Road & SW & Build & Wall & Fence & Pole & TL & TS & Veg. & Terrain & Sky & PR & Rider & Car & Truck & Bus & Train & Motor & Bike & mIoU  \\
\midrule
%------------------------------------ VGG GTA -> City ----------------------------------%
%---------------- Source Only ----------------
No Adapt & 75.8 & 16.8 & 77.2 & 12.5 & 21.0 & 25.5 & 30.1 & 20.1 & 81.3 & 24.6 & 70.3 & 53.8 & 26.4 & 49.9 & 17.2 & 25.9 & 6.5 & 25.3 & 36.0 & 36.6 \\
\midrule
%------------------------------------ RES GTA -> City ----------------------------------%
%---------------- Adaptseg ----------------%
AdaptSegNet~\cite{tsai2018learning} & 86.5 & 36.0 & 79.9 & 23.4 & 23.3 & 35.2 & 14.8 & 14.8 & 83.4 & 33.3 & 75.6 & 58.5 & 27.6 & 73.7 & 32.5 & 35.4 & 3.9 & 30.1 & 28.1 & 42.4 \\
%---------------- ADVENT ----------------%
ADVENT~\cite{vu2019advent} & 89.9 & 36.5 & 81.2 & 29.2 & 25.2 & 28.5 & 32.3 & 22.4 & 83.9 & 34.0 & 77.1 & 57.4 & 27.9 & 83.7 & 29.4 & 39.1 & 1.5 & 28.4 & 23.3 & 43.8\\
%---------------- Differential [CVPR 2020] ----------------%
SIMDA~\cite{wang2020differential} & 90.6 & 44.7 & 84.8 & 34.3 & 28.7 & 31.6 & 35.0 & 37.6 & 84.7 & 43.3 & 85.3 & 57.0 & 31.5 & 83.8 & 42.6 & 48.5 & 1.9 & 30.4 & 39.0 & 49.2\\
%---------------- Learning texture [CVPR 2020] ----------------%
LTIR~\cite{Kim2020LearningTI} & 92.9 & 55.0 & 85.3 & 34.2 & 31.1 & 34.9 & 40.7 & 34.0 & 85.2 & 40.1 & 87.1 & 61.0 & 31.1 & 82.5 & 32.3 & 42.9 & 0.3 & 36.4 & 46.1 & 50.2\\
%---------------- Phase DA [CVPR 2020] ----------------%
PCEDA~\cite{yang2020phase} & 91.0 & 49.1 & 85.6 & 37.2 & 29.7 & 33.7 & 38.1 & 39.2 & 85.4 & 35.4 & 85.1 & 61.1 & 32.8 & 84.1 & 45.6 & 46.9 & 0.0 & 34.2 & 44.5 & 50.5\\
%---------------- Fourier DA [CVPR 2020] ----------------%
FDA~\cite{yang2020fda} & 92.5 & 53.3 & 82.4 & 26.5 & 27.6 & 36.4 & 40.6 & 38.9 & 82.3 & 39.8 & 78.0 & 62.6 & 34.4 & 84.9 & 34.1 & 53.1 & 16.9 & 27.7 & 46.4 & 50.5 \\
%---------------- CBST ----------------%
CBST~\cite{zou2018unsupervised} & 91.8 & 53.5 & 80.5 & 32.7 & 21.0 & 34.0 & 28.9 & 20.4 & 83.9 & 34.2 & 80.9 & 53.1 & 24.0 & 82.7 & 30.3 & 35.9 & 16.0 & 25.9 & 42.8 & 45.9 \\
% %---------------- Ours( BT  + VOTE + IST ) ----------------%
% \rowcolor{lightgray} Ours+CBST & & 89.9 & 57.3 & 80.4 & 32.9 & 16.4 & 37.7 & 30.1 & 21.5 & 85.4 & 39.8 & 81.5 & 60.4 & 28.5 & 79.0 & 26.8 & 24.3 & \textbf{34.0} & 32.9 & 44.8 & 47.6 & 30.8  \\
%---------------- CRST ----------------%
CRST(MRKLD)~\cite{Zou_2019_ICCV} & 91.0 & 55.4 & 80.0 & 33.7 & 21.4 & 37.3 & 32.9 & 24.5 & 85.0 & 34.1 & 80.8 & 57.7 & 24.6 & 84.1 & 27.8 & 30.1 & 26.9 & 26.0 & 42.3 & 47.1 \\
TPLD~\cite{shin2020two} & 94.2 & 60.5 & 82.8 & 36.6 & 16.6 & 39.3 & 29.0 & 25.5 & 85.6 & 44.9 & 84.4 & 60.6 & 27.4 & 84.1 & 37.0 & 47.0 & 31.2 & 36.1 & 50.3 & 51.2   \\
IAST~\cite{mei2020instance}  & 93.8 & 57.8 & 85.1 & 39.5 & 26.7 & 26.2 & 43.1 & 34.7 & 84.9 & 32.9 & 88.0 & 62.6 & 29.0 & 87.3 & 39.2 & 49.6 & 23.2 & 34.7 & 39.6 & 51.5   \\
\midrule
WDA~\cite{Paul_WeakSegDA_ECCV20} (Point) & 94.0 & 62.7 & 86.3 & 36.5 & 32.8 & 38.4 & 44.9 & 51.0 & 86.1 & 43.4 & 87.7 & 66.4 & 36.5 & 87.9 & 44.1 & 58.8 & 23.2 & 35.6 & 55.9 & 56.4  \\
\midrule
\rowcolor{lightgray} Ours (PPL: Point) & 96.1 & 71.8 & 88.8 & 47.0 & 46.5 & 42.2 & 53.1 & 60.6 & 89.4 & 55.1 & 91.4 & 70.8 & 44.7 & 90.6 & 56.7 & 47.9 & 39.1 & 47.3 & 62.7 & \textbf{63.5}  \\
\rowcolor{lightgray} Ours (SPL: Segment) & \textbf{96.6} & \textbf{77.0} & \textbf{89.6} & \textbf{47.8} & \textbf{50.7} & \textbf{48.0} & \textbf{56.6} & \textbf{63.5} & \textbf{89.5} & \textbf{57.8} & \textbf{91.6} & \textbf{72.0} & \textbf{47.3} & \textbf{91.7} & \textbf{62.1} & \textbf{61.9} & \textbf{48.9} & \textbf{47.9} & \textbf{65.3} & \textbf{66.6}   \\
\midrule
Supervised & 96.9 & 77.1 & 89.8 & 45.6 & 49.9 & 47.4 & 55.8 & 64.1 & 90.0 & 58.2 & 92.8 & 71.9 & 46.9 & 91.4 & 60.3 & 65.8 & 54.3 & 44.6 & 64.7 & 66.7 \\
\bottomrule
\end{tabular}
}
\end{center}
\vspace{-1mm}
\caption{\textbf{Experimental results on GTA5 $\rightarrow$ Cityscapes.} 
While our PPL method already surpass previous UDA state-of-the-art models (e.g., IAST~\cite{mei2020instance}) and DA model with few labels(e.g., WDA~\cite{Paul_WeakSegDA_ECCV20}) by only leveraging (around 40 labeled points per image), our SPL method shows the performance comparable with fully supervised learning (only 0.1\% mIoU gap).}
\label{table:gta5_activ}
\end{table*}

\section{Experiments}

% (1) Ablation study on uncertainty measures (+random)  $\rightarrow$ O

% (2) Comparison with UDA (+supervised) SOTA $\rightarrow$ O

% (3) Label ratio VS performance Curve $\rightarrow$ O

% (4) Active Stage VS performance Curve $\rightarrow$ O

% (5) Ablation study on point selection approaches (ablation chart)

% (6) Qualitative Results (segmentation results) $\rightarrow$ O

% (7) Analysis (Visualizing generated masks for SPL, PPL)

% (8) Other analysis? 1. Efficiency, 2. Statistics

% (9) Combining with previous Pseudo label method (CBST, IAST)

% (10) 

In this section, we conduct extensive experiments to analyze our methods both quantitatively and qualitatively.
%We also evaluate our final adaptation results both quantitatively and qualitatively in comparison with the state-of-the-art baselines.

\begin{figure*}[t]
    \centering 
    \includegraphics[width=0.95\textwidth]{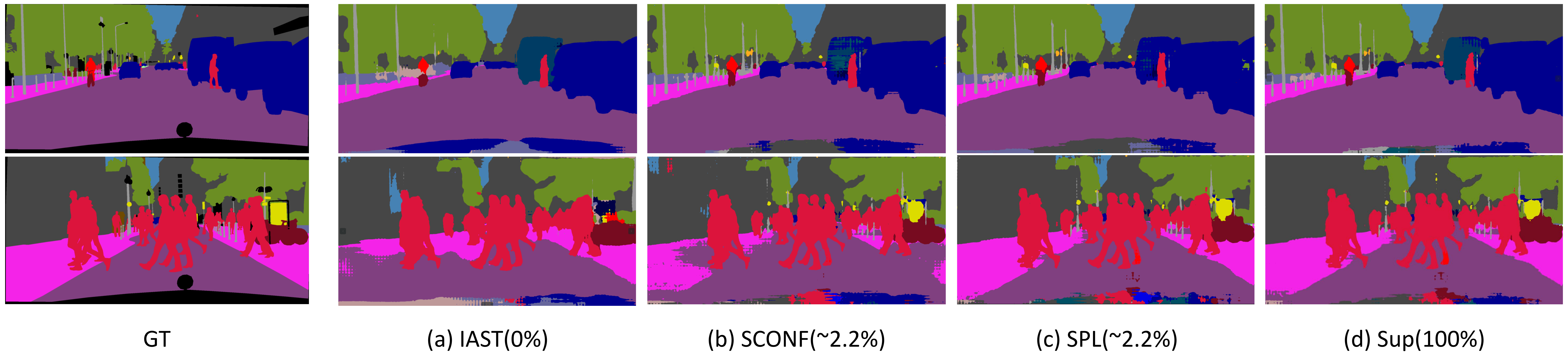}
    \caption{\textbf{Qualitative result of our SPL} 
    While the state-of-the-art UDA method, i.e., IAST~\cite{mei2020instance}, and a naive way to label regions, \textbf{SCONF} baseline, show erroneous segmentation results, the proposed method, SPL, shows the correct segmentation result similar to the fully supervised approach.}
    \label{fig:quality_spl}
    \vspace{-3mm}
\end{figure*}

\begin{figure*}[t]
    \centering 
    \includegraphics[width=0.95\textwidth]{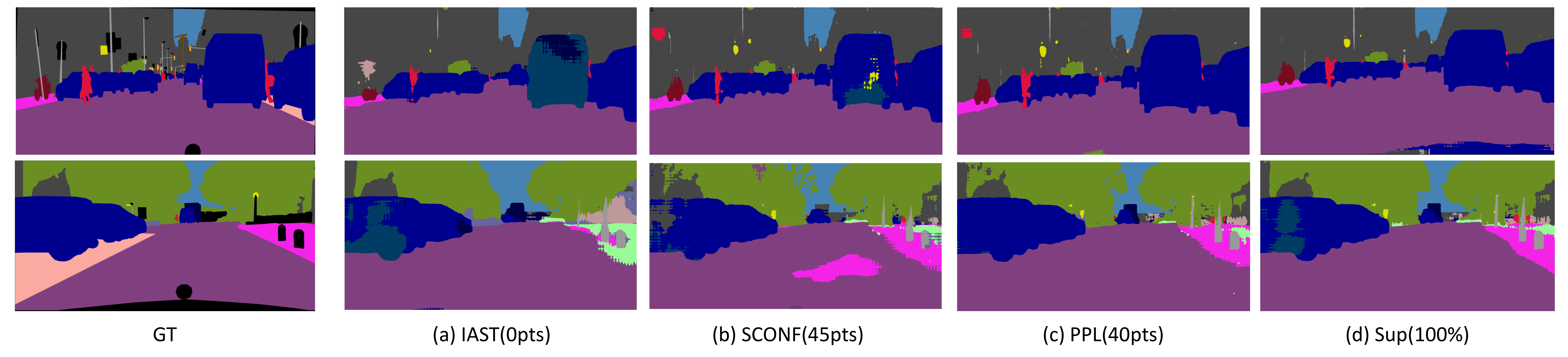}
    \caption{\textbf{Qualitative result of our PPL} While the state-of-the-art UDA method, i.e., IAST~\cite{mei2020instance}, and a naive way to la label regions, \textbf{SCONF} baseline, show erroneous segmentation results, the proposed method, PPL, shows the correct segmentation result similar to the fully supervised approach.}
    \label{fig:quality_ppl}
    \vspace{-3mm}
\end{figure*}

\subsection{Dataset}
We evaluate our model on the most common adaptation benchmark of GTA5~\cite{Richter_2016_ECCV} to Cityscapes~\cite{Cordts2016Cityscapes}.
% GTA5 contains 24966 synthetic images and Cityscapes contains ... training and ... evaluation images. 
Following the standard protocols from previous works~\cite{mei2020instance,Yawei2019Taking}, we adapt the model to the Cityscapes training set and evaluate the performance on the validation set.

\subsection{Implementation details}
To push the state-of-the-art benchmark performances, we test our method LabOR on the IAST framework~\cite{mei2020instance}.
For our backbones, we use ResNet-101~\cite{resnet} for the feature extractor and Deeplab-v2~\cite{deep2} for the segmentation model.
% We pretrain the model on ImageNet~\cite{imagenet} and fine-tune on the source domain images with SGD
We utilize source domain to pretrain model and adversarial training to initially reduce domain shift.
% For fair comparisons, we set the total self-training rounds to be same (8 rounds) with the previous approaches~\cite{zou2018unsupervised,Zou_2019_ICCV}
We train the model for a total of 3 stages. In each stage, the proposed iterative human-in-the-loop mechanism is performed.
We follow IAST's implementation details for fair comparison.
% Following training protocol in IAST~\cite{mei2020instance}, we randomly cropped and resized to $512 \times 1024$.
\subsection{Experimental Results on GTA5 $\to$ Cityscapes}
% Data loading follows the baseline protocols~\cite{Zhang_2017_ICCV,zou2018unsupervised}.\\

We show our quantitative results of both of our methods PPL and SPL compared to other state-of-the-art UDA methods~\cite{Yawei2019Taking,tsai2018learning,vu2019advent, zou2018unsupervised,Zou_2019_ICCV} in \tabref{table:gta5_activ}. Although out of our scope, we compare our method to Weak-label DA (WDA)~\cite{Paul_WeakSegDA_ECCV20} to show the competitiveness of our approach. To truly understand the capabilities of our approach, we also include the result of the fully supervised model.
\tabref{table:gta5_activ} shows that our LabOR SPL outperforms all state-of-the-art UDA or WDA approaches in all cases by a large margin. Even when compared to the fully supervised method, SPL is only down by 0.1 mIoU in comparison. In some classes such as ``Wall, Fence, Pole, TL, PR, Rider, Car, Truck, Motor, Bike,'' SPL even outperforms the supervised model. We believe this is a remarkable finding that can potentially be explored to hopefully surpass the performance of fully supervised methods. 
Even though our LabOR PPL only utilized point level supervision for the target dataset,
PPL also shows significant performance gains over previous state-of-the-art UDA or WDA methods. In comparison to the best performing UDA model IAST~\cite{mei2020instance}, PPL gains an 12\% increase in mIoU and the performance only degrades by 3.1\% when compared to SPL. Even when compared to WDA that utilizes point labels similar to PPL, our PPL has a 7.1\% increase in performance.
% which is a significant gap compared to the performance gaps among UDA methods (less than 3\% mIoU). %We believe this is a significant finding, 
Note that WDA labels average around 10~15 pixels per image, and although ours does give 3 times more pixels, we are able to increase the performance drastically while further reducing human interference.

%Clearly, we see that SPL is much more similar to the ground truth than the UDA plot. As for PPL, we see that the segmentation is slightly less clear in comparison to SPL, but still much closer to the ground truth that the UDA's output. 
%We can clearly see that even qualitatively both our methods have more visually pleasing results.

% For example, with Deeplab-v2 and ResNet-101 backbone, our TPLD significantly outperforms CRST by 4.2\%. 
% Moreover, to analyze the effect on rare classes, we also put rare-class mIoU.
% With the R-mIoU metric, we see the improvement is even much higher; 4.8\%.
% We provide qualitative results in Figure~\ref{fig:quality}. 
% Clearly, our final model generates the most visually pleasurable results.

\begin{figure*}[t]
    \centering 
    \includegraphics[width=0.95\textwidth]{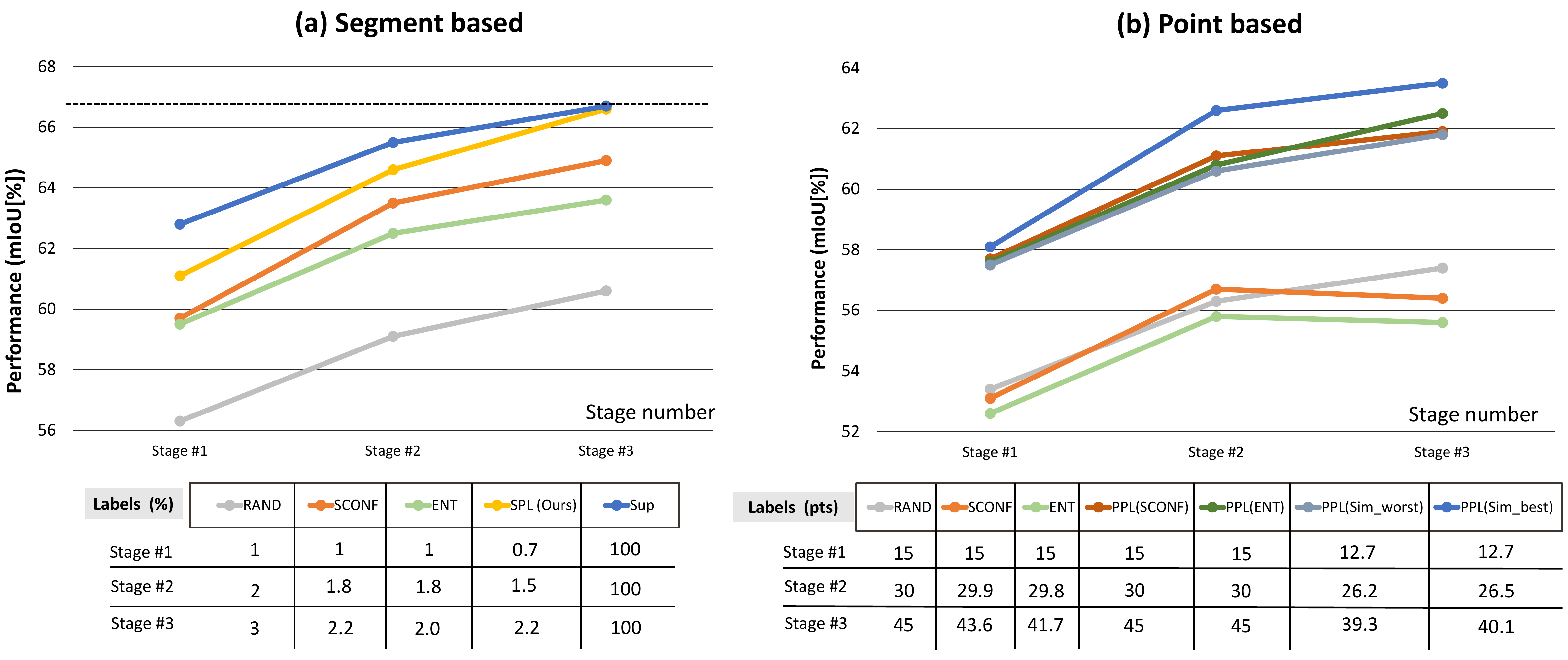}
    \caption{
    %\jw{(To DO: Do we need to include TE? change PPL(CONF) to PPL(SCONF))}
    \textbf{The performance of (a) segment based and (b) point based pixel labeling strategies.} (a) Our method, \textbf{SPL}, significantly outperforms all the methods among the uncertainty metrics, and our method shows the performance comparable to that of fully supervised training method at the final stage. (b) Among the point based strategies, our final model, (\textbf{PPL-Sim(best)}), shows the best performance.
    }
    \label{fig:loop_active}
    \vspace{-3mm}
\end{figure*}

% \subsection{Comparison against other pixel selector methods \djkim{(Analysis on Design Choices?)}}
\subsection{Further Discussion}

\begin{figure*}[t]
    \centering 
    \includegraphics[width=0.95\textwidth]{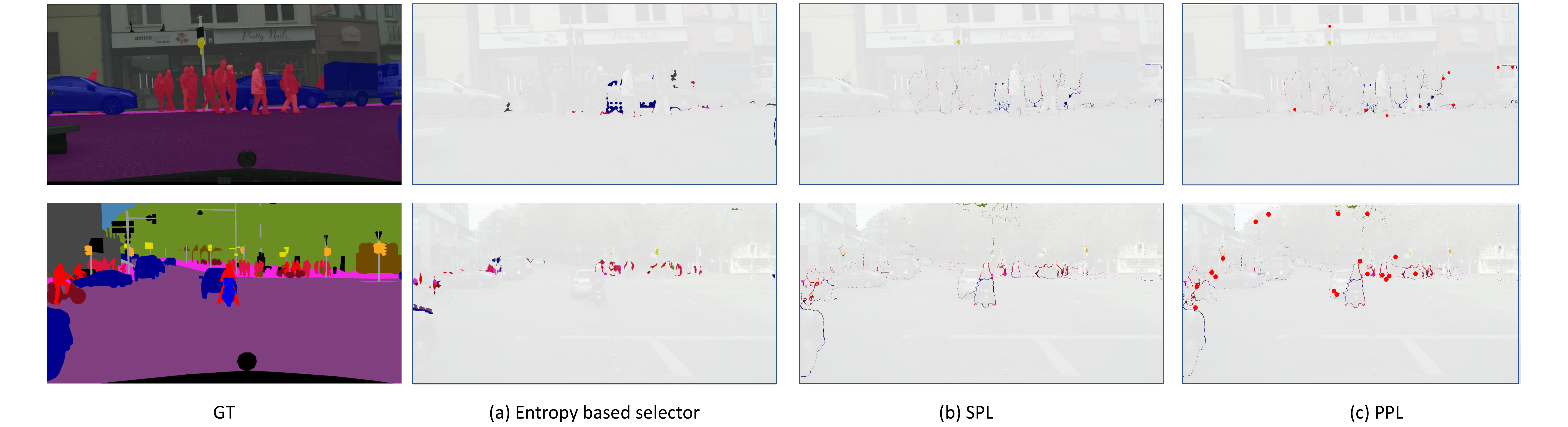}
    \caption{\textbf{The visualization of the generated regions to label} Compared to simple \textbf{ENT} baseline, our SPL and PPL are able to select more diverse points to give labels.}
    \label{fig:visualize}
    \vspace{-3mm}
\end{figure*}

%\noindent\textbf{SPL}
\noindent\textbf{Segment based Pixel-Labeling Strategies.}
To understand the performance gains from SPL as an uncertainty measure, we compare SPL with several other uncertainty metrics motivated from active learning research.
The comparison result between our \textbf{SPL} and the above baselines is demonstrated in \figref{fig:loop_active} (a).
\textbf{Random (RAND)} is a passive learning strategy that labels pixels according to a uniform distribution over an image region.
\textbf{Softmax Confidence (SCONF)}~\cite{culotta2005reducing} queries pixels for which a model has the least confidence in its most likely generated sequence: $1-\max_{K}\hat{\mathbf{Y}}_{i,j}$.
%, where $\hat{\mathbf{y}}^*$ is the probability of the most confident model's response.
\textbf{Entropy (ENT)}~\cite{shen2017deep} queries pixels that maximize the entropy of a model's output: $H(\hat{\mathbf{Y}}_{i,j})$, where $H(p) = -\sum_{k=1}^K p(k) \log p(k)$.
%We compare our methods by giving a constant number of pixel labels that are similar to our method. 
%At each stage, we give 1\% pixel labels and compare our methods against Random, Least Confident, and Entropy
For \textbf{RAND}, \textbf{SCONF}, and \textbf{ENT}, we need to set a constant number of pixels to label.
As a result, we give labels to 1\% pixel per image each stage for these baselines, so that the number of labeled pixels per stage is similar to that of our method.
Note that unlike \textbf{RAND}, the pixel selection of \textbf{SCONF} and \textbf{ENT} is dependent on the model's output during training, and this might cause the overlapping of some selected pixels over stages. As a result, although we give 1\% pixel labels per each stage, the accumulated number of labeled pixels might be lower than 1\%$\times$(Stage) as shown in \figref{fig:loop_active} (a).
% In addition, a more sophisticated uncertainty metric, \textbf{Temporal Ensembling (TE)}, is similar to our inconsistency mask, but it is to compare the outputs our a model along different epochs:
% \begin{equation}
%     M(\mathbf{x}_t) = \bigl[ \argmax_K f_{\theta^{(t)}} \circ g_{\phi^{(t)}} (\mathbf{x}_t)  \neq \argmax_K f_{\theta^{(t+1)}} \circ g_{\phi^{(t+1)}} (\mathbf{x}_t) \bigr].
% \end{equation}
\textbf{Fully Supervised (Sup)} leverages the all the ground truth labels in the target dataset for training. 
%Using these metrics, we compare to SPL how each method of giving labels affect the performance. 
As shown in \figref{fig:loop_active} (a), \textbf{SPL} significantly outperforms \textbf{SCONF}, which is the best performing method among the uncertainty metrics. 
In addition, even though \textbf{Sup} shows 1.67\% mIoU gap in relation to \textbf{SPL} at Stage 1,  our method shows only a 0.1\% mIoU gap with \textbf{Sup} at the final stage (Stage 3).
Furthermore, we tested the supervised baseline and our SPL to Stage 4, but the performances of both models show the same as that of Stage 3, therefore, we make a decision to only train all methods only up to Stage 3. 
In summary, our \textbf{SPL} is the best option among various possible uncertain pixel selection methods in terms of the performance gain.

%\noindent\textbf{PPL}
\noindent\textbf{Point based Pixel-Labeling Strategies.}
For PPL, there are various options to select the \emph{pixels} to label. 
We perform an additional experiment comparing our PPL with other several approaches in \figref{fig:loop_active} (b).
In addition to our PPL distance measures, we also evaluate other pixel selection methods,
\textbf{RAND}, \textbf{SCONF}, and \textbf{ENT}, that are the exact same uncertainty metrics described in the previous paragraph, but we give labels to 15 pixels per image each stage for these baselines this time, so that the number of labeled pixels per stage is similar to that of our method.
Given an inconsistency mask from \eqnref{eq:inconsistency}, there are various options to select the representative points among the pixels other than measuring the distance with the class prototypes. 
\textbf{PPL-SCONF} queries pixels among the inconsistency mask for which a model has the least confidence in its most likely generated sequence.
\textbf{PPL-ENT} queries pixels among the inconsistency mask that maximize the entropy of a model's output.
Note that once again for \textbf{SCONF} and \textbf{ENT}, some uncertain pixels are overlapped, causing the number of pixels to be less than 15 each stage.
After we measure the distance between the prototype vectors and the output prediction for the pixels, we can either select the point that is the nearest (\textbf{PPL-Sim(best)}) or far (\textbf{PPL-Sim(worst)}) from the prototype vectors.
\figref{fig:loop_active} (b) shows that our final PPL model, (\textbf{PPL-Sim(best)}), shows the best performance.
Note that even the worst PPL distance measure of \textbf{PPL-Sim(worst)} far outperforms any of the other non-PPL based methods by a large margin. 
Interestingly, although \textbf{RAND} performs the best among the non-PPL based methods at Stage 3, \textbf{PPL-Sim(best)} even at Stage 1 outperforms the best performance of \textbf{RAND}.
This result shows the importance of the strategy to pick pixels to label for a model's performance.
% Although it is natural that giving some labels does increase the performance of a model, we conclude that giving labels to the uncertain pixels improves the performance the most.

%Although in our final PPL model use cosine similarity as in \eqnref{eq:ppleq}, 
%In addition to PPL distance measures, we also test other pixel selection methods such as Random (uniform), Least Confident, and Entropy to show the performance gains of PPL. 
%Interestingly, among the non-PPL based methods, Random performs the best at Stage 3, but even at Stage 1, PPL(Sim-best)\jw{(maybe we can change name of this)} still outperforms it. 
%The results here shows how dependent a model is on which points to pick. 
%Naturally giving some labels will definitely increase performance as we would transition from unsupervised to some-supervised, but we show that giving the desired (or uncertain) pixels drastically improve performance.

\noindent\textbf{Qualitative Results.}
\figref{fig:quality_spl} and \figref{fig:quality_ppl} show the qualitative results of both our methods, SPL and PPL respectively, in comparison to the ground truth, the state-of-the-art UDA method, IAST~\cite{mei2020instance}, \textbf{SCONF} baseline for uncertain region selection, and \textbf{Sup} baseline as a performance upper bound. 
In \figref{fig:quality_spl}, while IAST and \textbf{SCONF} baseline show erroneous segmentation results (e.g., the class ``car'' in the top result and the class ``sidewalk'' in the bottom result), the proposed method, SPL, shows the correct segmentation result similar to the supervised approach.
In \figref{fig:quality_ppl}, IAST confuses the class ``car'' as ``bus'' and fails to classify the class ``sidewalk.'' \textbf{SCONF} baseline generates noisy segmentation result.
In contrast, the proposed method, PPL, shows the correct segmentation result similar to the supervised approach.

\figref{fig:visualize} visualizes the selected uncertain pixels to label from the \textbf{ENT} baseline and our methods SPL and PPL.
We can see that unlike \textbf{ENT}, SPL is able to cover a much wider range of pixels across the image. \textbf{ENT} on the other hand tends to lump pixels that are nearby together. Futhermore, PPL is also shown to pick diverse pixels and not be grounded to a certain region of the image.

\noindent\textbf{Effects of Entropy Regularization on SPL and PPL.}
Recent work~\cite{mei2020instance} has proposed a regularizer in the form of entropy minimization for training in UDA to regularize uncertain points in an image. In light of this, we apply the entropy minimizer on both SPL and PPL to test its effect on performance. \tabref{tab:Abl_reg} shows the effects of adding the entropy minimizer. Interestingly, on SPL, the entropy minimizer does not seem to have much impact. At Stages 1 and 2, the performance does increase slightly, but at Stage 3, the performance decreases. In contrast, for PPL, the entropy regularizer slightly improves the performance. We believe this might be the case as for SPL as the uncertain pixels of SPL are given ground truth labels for, so the regularizer has minimal effects. For PPL, as the number of ground truth pixels given are few, the regularizer helps in model training.

% %%%%%%%%%%%%%%%%%%%%%%%%%%%%%%%%%%%%%%%%%%%%%%%%%%%%%%%%%%%
% \begin{table}[t]
% \centering
% \resizebox{0.48\textwidth}{!}
% {
% \def\arraystretch{1.1}
% \begin{tabular}{lccccc}
% \hline
% % \multicolumn{1}{c}{Method} & \multicolumn{1}{c}{Discover} & \multicolumn{1}{c}{Hallucinate} & \multicolumn{1}{c}{Adapt} & \multicolumn{1}{c}{C}& & \multicolumn{1}{c}{C+O} \\
% % \multicolumn{1}{c}{} & \multicolumn{2}{c}{Dice} & \multicolumn{2}{c}{Adv.} & \multicolumn{2}{c}{Metric}  \\
% Method & Pseudo Label & Stage\#1 & Stage\#2 & Stage\#3 \\
% \hline
% \hline
% % Source  &       &        &      &       & 34.0  & 21.2 \\
% \multirow{3}{*}{SPL}  &  No cut  &  61.1\%(0.7\%)  &  64.6\%(1.5\%)  & \textbf{66.6\%(2.2\%)}  \\
%  & CBST~\cite{zou2018domain}  &   60.9\%(3.1\%)    & 64.6\%(6.3\%)   & 65.3\%(8.3\%)  \\
%  & IAST~\cite{mei2020instance}  &   60.5\%(2.6\%)    & 64.1\%(5.6\%)   & 65.2\%(8.5\%)    \\
% \hline
% \end{tabular}
% }
% \caption{\textbf{Effect of pseudo label generation on SPL accuracy and label ratio.} The existing pseudo label thresholding techniques from CBST~\cite{Zou_2019_ICCV} and IAST~\cite{mei2020instance} do not improve the performance of our SPL.}
% % \vspace{1mm}
% \label{tab:Abl_cut}
% \end{table}

\begin{table}[t]
\centering
\resizebox{0.48\textwidth}{!}
{
\def\arraystretch{1.1}
\begin{tabular}{lccccc}
\hline
% \multicolumn{1}{c}{Method} & \multicolumn{1}{c}{Discover} & \multicolumn{1}{c}{Hallucinate} & \multicolumn{1}{c}{Adapt} & \multicolumn{1}{c}{C}& & \multicolumn{1}{c}{C+O} \\
% \multicolumn{1}{c}{} & \multicolumn{2}{c}{Dice} & \multicolumn{2}{c}{Adv.} & \multicolumn{2}{c}{Metric}  \\
Method & Regularizer & Stage\#1 & Stage\#2 & Stage\#3 \\
\hline
\hline
% Source  &       &        &      &       & 34.0  & 21.2 \\
\multirow{2}{*}{SPL}  &  $\times$  &  61.1\%(0.7\%)  &  64.6\%(1.5\%)  & \textbf{66.6\%(2.2\%)} \\
 & Ent~\cite{mei2020instance}  &  61.5\%(0.7\%)    & 64.9\%(1.4\%)   & 66.4\%(2.1\%) \\
 \hline
 \multirow{2}{*}{PPL}& $\times$  &   58.1\%(12.7pts)    & 62.6\%(26.5pts)   & 63.5\%(40.1pts)     \\
 & Ent~\cite{mei2020instance}  &   58.9\%(12.7pts)    & 62.3\%(26.3pts)   & 63.9\%(39.4pts)    \\
\hline
\end{tabular}
}
\vspace{1mm}
\caption{\textbf{Effect of self-training entropy regularization~\cite{mei2020instance} on SPL and PPL.} While the entropy regularization does not improve the performance of our SPL, adding entropy regularize on our PPL slightly improves the performance.}
\vspace{-3mm}
\label{tab:Abl_reg}
\end{table}

% Conclusion
\section{Conclusion}

In this work, we tackle performance discrepancy of Unsupervised Domain Adaptation and proposed a new framework for domain adaptive semantic segmentation in a human-in-the-loop manner while generating the most informative pixel points that we call \textbf{Lab}eling \textbf{O}nly if \textbf{R}equired, \textbf{LabOR}. Based on a self-training platform, we build our method to select the most \emph{informative} pixels and introduce two pixel selection methods that we call ``Segment based Pixel-Labeling'' and ``Point based Pixel-Labeling.'' Through our experiments, we demonstrate the effectiveness of our approach and show near supervised performance while drastically lowering human annotation costs. 
We believe that our work opens a new paradigm of domain adaptation and challenge future research to be performed in this area to hopefully surpass the fully supervised method.

\ificcvfinal\thispagestyle{empty}\fi

{\small
\bibliographystyle{ieee_fullname}
\bibliography{egbib}

\begin{thebibliography}{10}\itemsep=-1pt

\bibitem{atapour2018real}
Amir Atapour-Abarghouei and Toby~P Breckon.
\newblock Real-time monocular depth estimation using synthetic data with domain
  adaptation via image style transfer.
\newblock In {\em Proceedings of the IEEE Conference on Computer Vision and
  Pattern Recognition}, pages 2800--2810, 2018.

\bibitem{deep2}
Liang-Chieh Chen, George Papandreou, Iasonas Kokkinos, Kevin Murphy, and Alan~L
  Yuille.
\newblock Deeplab: Semantic image segmentation with deep convolutional nets,
  atrous convolution, and fully connected crfs.
\newblock {\em IEEE transactions on pattern analysis and machine intelligence},
  40(4):834--848, 2017.

\bibitem{deep3}
Liang-Chieh Chen, George Papandreou, Florian Schroff, and Hartwig Adam.
\newblock Rethinking atrous convolution for semantic image segmentation.
\newblock {\em arXiv preprint arXiv:1706.05587}, 2017.

\bibitem{chen2017show}
Tseng-Hung Chen, Yuan-Hong Liao, Ching-Yao Chuang, Wan-Ting Hsu, Jianlong Fu,
  and Min Sun.
\newblock Show, adapt and tell: Adversarial training of cross-domain image
  captioner.
\newblock In {\em {Proc. of Int'l Conf. on Computer Vision (ICCV)}}, 2017.

\bibitem{chen2018domain}
Yuhua Chen, Wen Li, Christos Sakaridis, Dengxin Dai, and Luc Van~Gool.
\newblock Domain adaptive faster r-cnn for object detection in the wild.
\newblock In {\em {Proc. of Computer Vision and Pattern Recognition (CVPR)}},
  pages 3339--3348, 2018.

\bibitem{Chen_2019_CVPR}
Yun-Chun Chen, Yen-Yu Lin, Ming-Hsuan Yang, and Jia-Bin Huang.
\newblock Crdoco: Pixel-level domain transfer with cross-domain consistency.
\newblock In {\em {Proc. of Computer Vision and Pattern Recognition (CVPR)}},
  June 2019.

\bibitem{cho2021mcdal}
Jae~Won Cho, Dong-Jin Kim, Yunjae Jung, and In~So Kweon.
\newblock Mcdal: Maximum classifier discrepancy for active learning.
\newblock {\em arXiv preprint arXiv:2107.11049}, 2021.

\bibitem{Cordts2016Cityscapes}
Marius Cordts, Mohamed Omran, Sebastian Ramos, Timo Rehfeld, Markus Enzweiler,
  Rodrigo Benenson, Uwe Franke, Stefan Roth, and Bernt Schiele.
\newblock The cityscapes dataset for semantic urban scene understanding.
\newblock In {\em Proceedings of the IEEE conference on computer vision and
  pattern recognition}, pages 3213--3223, 2016.

\bibitem{culotta2005reducing}
Aron Culotta and Andrew McCallum.
\newblock Reducing labeling effort for structured prediction tasks.
\newblock In {\em {Proc. of Association for the Advancement of Artificial
  Intelligence (AAAI)}}, 2005.

\bibitem{fernando2013unsupervised}
Basura Fernando, Amaury Habrard, Marc Sebban, and Tinne Tuytelaars.
\newblock Unsupervised visual domain adaptation using subspace alignment.
\newblock In {\em {Proc. of Int'l Conf. on Computer Vision (ICCV)}}, pages
  2960--2967, 2013.

\bibitem{ganin2014unsupervised}
Yaroslav Ganin and Victor Lempitsky.
\newblock Unsupervised domain adaptation by backpropagation.
\newblock {\em arXiv preprint arXiv:1409.7495}, 2014.

\bibitem{ghifary2016deep}
Muhammad Ghifary, W~Bastiaan Kleijn, Mengjie Zhang, David Balduzzi, and Wen Li.
\newblock Deep reconstruction-classification networks for unsupervised domain
  adaptation.
\newblock In {\em {Proc. of European Conf. on Computer Vision (ECCV)}}, pages
  597--613. Springer, 2016.

\bibitem{sim2robot}
Florian Golemo, Adrien~Ali Taiga, Aaron Courville, and Pierre-Yves Oudeyer.
\newblock Sim-to-real transfer with neural-augmented robot simulation.
\newblock In Aude Billard, Anca Dragan, Jan Peters, and Jun Morimoto, editors,
  {\em Proceedings of The 2nd Conference on Robot Learning}, volume~87 of {\em
  Proceedings of Machine Learning Research}, pages 817--828. PMLR, 29--31 Oct
  2018.

\bibitem{gong2012geodesic}
Boqing Gong, Yuan Shi, Fei Sha, and Kristen Grauman.
\newblock Geodesic flow kernel for unsupervised domain adaptation.
\newblock In {\em {Proc. of Computer Vision and Pattern Recognition (CVPR)}},
  pages 2066--2073. IEEE, 2012.

\bibitem{gopalan2011domain}
Raghuraman Gopalan, Ruonan Li, and Rama Chellappa.
\newblock Domain adaptation for object recognition: An unsupervised approach.
\newblock In {\em 2011 international conference on computer vision}, pages
  999--1006. IEEE, 2011.

\bibitem{resnet}
Kaiming He, Xiangyu Zhang, Shaoqing Ren, and Jian Sun.
\newblock Deep residual learning for image recognition.
\newblock In {\em Proceedings of the IEEE conference on computer vision and
  pattern recognition}, pages 770--778, 2016.

\bibitem{pmlr-v80-hoffman18a}
Judy Hoffman, Eric Tzeng, Taesung Park, Jun-Yan Zhu, Phillip Isola, Kate
  Saenko, Alexei Efros, and Trevor Darrell.
\newblock {C}y{CADA}: Cycle-consistent adversarial domain adaptation.
\newblock In {\em {Proc. of Int'l Conf. on Machine Learning (ICML)}}, pages
  1989--1998, 2018.

\bibitem{fcnwild}
Judy Hoffman, Dequan Wang, Fisher Yu, and Trevor Darrell.
\newblock Fcns in the wild: Pixel-level adversarial and constraint-based
  adaptation.
\newblock {\em arXiv preprint arXiv:1612.02649}, 2016.

\bibitem{Hong_2018_CVPR}
Weixiang Hong, Zhenzhen Wang, Ming Yang, and Junsong Yuan.
\newblock Conditional generative adversarial network for structured domain
  adaptation.
\newblock In {\em {Proc. of Computer Vision and Pattern Recognition (CVPR)}},
  June 2018.

\bibitem{kim2019image}
Dong-Jin Kim, Jinsoo Choi, Tae-Hyun Oh, and In~So Kweon.
\newblock Image captioning with very scarce supervised data: Adversarial
  semi-supervised learning approach.
\newblock In {\em Proceedings of the 2019 Conference on Empirical Methods in
  Natural Language Processing and the 9th International Joint Conference on
  Natural Language Processing (EMNLP-IJCNLP)}, 2019.

\bibitem{kim2018disjoint}
Dong-Jin Kim, Jinsoo Choi, Tae-Hyun Oh, Youngjin Yoon, and In~So Kweon.
\newblock Disjoint multi-task learning between heterogeneous human-centric
  tasks.
\newblock In {\em {Proc. of Winter Conference on Applications of Computer
  Vision (WACV)}}, 2018.

\bibitem{kim2020detecting}
Dong-Jin Kim, Xiao Sun, Jinsoo Choi, Stephen Lin, and In~So Kweon.
\newblock Detecting human-object interactions with action co-occurrence priors.
\newblock In {\em {Proc. of European Conf. on Computer Vision (ECCV)}}, 2020.

\bibitem{Kim2020LearningTI}
Myeongjin Kim and Hyeran Byun.
\newblock Learning texture invariant representation for domain adaptation of
  semantic segmentation.
\newblock In {\em Proceedings of the IEEE/CVF Conference on Computer Vision and
  Pattern Recognition}, pages 12975--12984, 2020.

\bibitem{kulis2011you}
Brian Kulis, Kate Saenko, and Trevor Darrell.
\newblock What you saw is not what you get: Domain adaptation using asymmetric
  kernel transforms.
\newblock In {\em {Proc. of Computer Vision and Pattern Recognition (CVPR)}},
  pages 1785--1792. IEEE, 2011.

\bibitem{li2017deeper}
Da Li, Yongxin Yang, Yi-Zhe Song, and Timothy~M Hospedales.
\newblock Deeper, broader and artier domain generalization.
\newblock In {\em {Proc. of Int'l Conf. on Computer Vision (ICCV)}}, pages
  5542--5550, 2017.

\bibitem{li2017domain}
Wen Li, Zheng Xu, Dong Xu, Dengxin Dai, and Luc Van~Gool.
\newblock Domain generalization and adaptation using low rank exemplar svms.
\newblock In {\em {IEEE Trans. Pattern Anal. Mach. Intell. (TPAMI)}},
  volume~40, pages 1114--1127. IEEE, 2017.

\bibitem{li2019bidirectional}
Yunsheng Li, Lu Yuan, and Nuno Vasconcelos.
\newblock Bidirectional learning for domain adaptation of semantic
  segmentation.
\newblock In {\em Proceedings of the IEEE/CVF Conference on Computer Vision and
  Pattern Recognition}, pages 6936--6945, 2019.

\bibitem{trasdeepadat}
Mingsheng Long, Yue Cao, Zhangjie Cao, Jianmin Wang, and Michael~I Jordan.
\newblock Transferable representation learning with deep adaptation networks.
\newblock {\em {IEEE Trans. Pattern Anal. Mach. Intell. (TPAMI)}},
  41(12):3071--3085, 2018.

\bibitem{long2015learning}
Mingsheng Long, Yue Cao, Jianmin Wang, and Michael~I Jordan.
\newblock Learning transferable features with deep adaptation networks.
\newblock {\em arXiv preprint arXiv:1502.02791}, 2015.

\bibitem{Yawei2019Taking}
Yawei Luo, Liang Zheng, Tao Guan, Junqing Yu, and Yi Yang.
\newblock Taking a closer look at domain shift: Category-level adversaries for
  semantics consistent domain adaptation.
\newblock In {\em {Proc. of Computer Vision and Pattern Recognition (CVPR)}},
  2019.

\bibitem{mei2020instance}
Ke Mei, Chuang Zhu, Jiaqi Zou, and Shanghang Zhang.
\newblock Instance adaptive self-training for unsupervised domain adaptation.
\newblock {\em arXiv preprint arXiv:2008.12197}, 2020.

\bibitem{motiian2017unified}
Saeid Motiian, Marco Piccirilli, Donald~A Adjeroh, and Gianfranco Doretto.
\newblock Unified deep supervised domain adaptation and generalization.
\newblock In {\em {Proc. of Int'l Conf. on Computer Vision (ICCV)}}, pages
  5715--5725, 2017.

\bibitem{imagetrans}
Z. {Murez}, S. {Kolouri}, D. {Kriegman}, R. {Ramamoorthi}, and K. {Kim}.
\newblock Image to image translation for domain adaptation.
\newblock In {\em {Proc. of Computer Vision and Pattern Recognition (CVPR)}},
  pages 4500--4509, June 2018.

\bibitem{pan2020unsupervised}
Fei Pan, Inkyu Shin, Francois Rameau, Seokju Lee, and In~So Kweon.
\newblock Unsupervised intra-domain adaptation for semantic segmentation
  through self-supervision.
\newblock In {\em Proceedings of the IEEE/CVF Conference on Computer Vision and
  Pattern Recognition}, pages 3764--3773, 2020.

\bibitem{panareda2017open}
Pau Panareda~Busto and Juergen Gall.
\newblock Open set domain adaptation.
\newblock In {\em {Proc. of Int'l Conf. on Computer Vision (ICCV)}}, pages
  754--763, 2017.

\bibitem{Paul_WeakSegDA_ECCV20}
Sujoy Paul, Yi-Hsuan Tsai, Samuel Schulter, Amit~K. Roy-Chowdhury, and Manmohan
  Chandraker.
\newblock Domain adaptive semantic segmentation using weak labels.
\newblock In {\em European Conference on Computer Vision (ECCV)}, 2020.

\bibitem{prabhu2020active}
Viraj Prabhu, Arjun Chandrasekaran, Kate Saenko, and Judy Hoffman.
\newblock Active domain adaptation via clustering uncertainty-weighted
  embeddings, 2020.

\bibitem{richter2017playing}
Stephan~R Richter, Zeeshan Hayder, and Vladlen Koltun.
\newblock Playing for benchmarks.
\newblock In {\em Proceedings of the IEEE International Conference on Computer
  Vision}, pages 2213--2222, 2017.

\bibitem{Richter_2016_ECCV}
Stephan~R. Richter, Vibhav Vineet, Stefan Roth, and Vladlen Koltun.
\newblock Playing for data: {G}round truth from computer games.
\newblock In Bastian Leibe, Jiri Matas, Nicu Sebe, and Max Welling, editors,
  {\em {Proc. of European Conf. on Computer Vision (ECCV)}}, volume 9906 of
  {\em LNCS}, pages 102--118. Springer International Publishing, 2016.

\bibitem{saito2019semisupervised}
Kuniaki Saito, Donghyun Kim, Stan Sclaroff, Trevor Darrell, and Kate Saenko.
\newblock Semi-supervised domain adaptation via minimax entropy, 2019.

\bibitem{saito2018maximum}
Kuniaki Saito, Kohei Watanabe, Yoshitaka Ushiku, and Tatsuya Harada.
\newblock Maximum classifier discrepancy for unsupervised domain adaptation.
\newblock In {\em Proceedings of the IEEE conference on computer vision and
  pattern recognition}, pages 3723--3732, 2018.

\bibitem{sener2016learning}
Ozan Sener, Hyun~Oh Song, Ashutosh Saxena, and Silvio Savarese.
\newblock Learning transferrable representations for unsupervised domain
  adaptation.
\newblock In {\em {Proc. of Neural Information Processing Systems (NeurIPS)}},
  pages 2110--2118, 2016.

\bibitem{settles2012active}
Burr Settles.
\newblock Active learning.
\newblock {\em Synthesis lectures on artificial intelligence and machine
  learning}, 6(1):1--114, 2012.

\bibitem{shen2017deep}
Yanyao Shen, Hyokun Yun, Zachary~C Lipton, Yakov Kronrod, and Animashree
  Anandkumar.
\newblock Deep active learning for named entity recognition.
\newblock {\em arXiv preprint arXiv:1707.05928}, 2017.

\bibitem{shin2020two}
Inkyu Shin, Sanghyun Woo, Fei Pan, and In~So Kweon.
\newblock Two-phase pseudo label densification for self-training based domain
  adaptation.
\newblock In {\em European Conference on Computer Vision}, pages 532--548.
  Springer, 2020.

\bibitem{learnSi}
Ashish Shrivastava, Tomas Pfister, Oncel Tuzel, Joshua Susskind, Wenda Wang,
  and Russell Webb.
\newblock Learning from simulated and unsupervised images through adversarial
  training.
\newblock In {\em {Proc. of Computer Vision and Pattern Recognition (CVPR)}},
  pages 2107--2116, 2017.

\bibitem{sohn2020fixmatch}
Kihyuk Sohn, David Berthelot, Chun{-}Liang Li, Zizhao Zhang, Nicholas Carlini,
  Ekin~D Cubuk, Alex Kurakin, Han Zhang, and Colin Raffel.
\newblock Fixmatch: Simplifying semi-supervised learning with consistency and
  confidence.
\newblock 2020.

\bibitem{su2020active}
Jong-Chyi Su, Yi-Hsuan Tsai, Kihyuk Sohn, Buyu Liu, Subhransu Maji, and
  Manmohan Chandraker.
\newblock Active adversarial domain adaptation, 2020.

\bibitem{tsai2018learning}
Yi-Hsuan Tsai, Wei-Chih Hung, Samuel Schulter, Kihyuk Sohn, Ming-Hsuan Yang,
  and Manmohan Chandraker.
\newblock Learning to adapt structured output space for semantic segmentation.
\newblock In {\em {Proc. of Computer Vision and Pattern Recognition (CVPR)}},
  pages 7472--7481, 2018.

\bibitem{vu2019advent}
Tuan-Hung Vu, Himalaya Jain, Maxime Bucher, Matthieu Cord, and Patrick
  P{\'e}rez.
\newblock Advent: Adversarial entropy minimization for domain adaptation in
  semantic segmentation.
\newblock In {\em Proceedings of the IEEE/CVF Conference on Computer Vision and
  Pattern Recognition}, pages 2517--2526, 2019.

\bibitem{wang2020alleviating}
Zhonghao Wang, Yunchao Wei, Rogerior Feris, Jinjun Xiong, Wen-Mei Hwu,
  Thomas~S. Huang, and Humphrey Shi.
\newblock Alleviating semantic-level shift: A semi-supervised domain adaptation
  method for semantic segmentation, 2020.

\bibitem{wang2020differential}
Zhonghao Wang, Mo Yu, Yunchao Wei, Rogerio Feris, Jinjun Xiong, Wen mei Hwu,
  Thomas~S. Huang, and Humphrey Shi.
\newblock Differential treatment for stuff and things: A simple unsupervised
  domain adaptation method for semantic segmentation, 2020.

\bibitem{yang2020phase}
Yanchao Yang, Dong Lao, Ganesh Sundaramoorthi, and Stefano Soatto.
\newblock Phase consistent ecological domain adaptation.
\newblock In {\em Proceedings of the IEEE/CVF Conference on Computer Vision and
  Pattern Recognition}, pages 9011--9020, 2020.

\bibitem{yang2020fda}
Yanchao Yang and Stefano Soatto.
\newblock Fda: Fourier domain adaptation for semantic segmentation.
\newblock In {\em Proceedings of the IEEE/CVF Conference on Computer Vision and
  Pattern Recognition}, pages 4085--4095, 2020.

\bibitem{zou2018domain}
Yang Zou, Zhiding Yu, BVK Kumar, and Jinsong Wang.
\newblock Domain adaptation for semantic segmentation via class-balanced
  self-training.
\newblock {\em arXiv preprint arXiv:1810.07911}, 2018.

\bibitem{zou2018unsupervised}
Yang Zou, Zhiding Yu, BVK~Vijaya Kumar, and Jinsong Wang.
\newblock Unsupervised domain adaptation for semantic segmentation via
  class-balanced self-training.
\newblock In {\em {Proc. of European Conf. on Computer Vision (ECCV)}}, pages
  289--305, 2018.

\bibitem{zou2019confidence}
Yang Zou, Zhiding Yu, Xiaofeng Liu, BVK Kumar, and Jinsong Wang.
\newblock Confidence regularized self-training.
\newblock In {\em Proceedings of the IEEE/CVF International Conference on
  Computer Vision}, pages 5982--5991, 2019.

\bibitem{Zou_2019_ICCV}
Yang Zou, Zhiding Yu, Xiaofeng Liu, B.V.K.~Vijaya Kumar, and Jinsong Wang.
\newblock Confidence regularized self-training.
\newblock In {\em The IEEE International Conference on Computer Vision (ICCV)},
  October 2019.

\end{thebibliography}
}

\end{document}